\documentclass[sigconf,nonacm]{acmart}

\AtBeginDocument{%
  \providecommand\BibTeX{{%
    \normalfont B\kern-0.5em{\scshape i\kern-0.25em b}\kern-0.8em\TeX}}}

\copyrightyear{2022}
\acmYear{2022}
\setcopyright{rightsretained}
\acmConference[GECCO '22 Companion]{Genetic and Evolutionary
Computation Conference Companion}{July 9--13, 2022}{Boston, MA, USA}
\acmBooktitle{Genetic and Evolutionary Computation Conference
Companion (GECCO '22 Companion), July 9--13, 2022, Boston, MA, USA}
\acmDOI{10.1145/3520304.3534003}
\acmISBN{978-1-4503-9268-6/22/07}

\ifdefined\N                                                                
\renewcommand{\N}{\mathds{N}}                                                
\else
  \newcommand{\N}{\mathds{N}}
\fi
\newcommand{\R}{\mathds{R}}                                                 
\ifdefined\C
  \renewcommand{\C}{\mathds{C}}                                             
\else
  \newcommand{\C}{\mathds{C}}
\fi

\DeclareMathOperator*{\argmin}{arg\,min}

\newcommand{\allDatasets}{\mathds{D}}                                       
\newcommand{\D}{\mathcal{D}}                                                      
\renewcommand{\xi}[1][i]{\mathbf{x}^{(#1)}}                                          

\newcommand{\preimageInducerShort}{\allDatasets\times\Lambda}     
\newcommand{\inducer}{\mathcal{I}}                                                

\newcommand{\Hspace}{\mathcal{H}}														
\newcommand{\fh}{\hat{f}}                                                   
\newcommand{\GEh}{\widehat{\mathrm{GE}}}                                             
           
\newcommand{\GEhlam}{\GEh(\lambdav)}                                                 

\newcommand{\lambdav}{\bm{\lambda}}											

\newcommand{\Ilam}{\inducer_{\lambdav}}						
\newcommand{\lams}{\lambdav^{*}}		                    
\newcommand{\LamS}{\tilde\Lambda}                           

\newtheorem{define}{Definition}

\usepackage{bm}
\usepackage{booktabs}
\usepackage{mathtools}
\usepackage{amsfonts}
\usepackage{dsfont}
\usepackage{cleveref}
\usepackage[flushleft]{threeparttable}
\usepackage{graphicx}
\usepackage{comment}
\usepackage{hyperref}

\begin{document}

\title[A Collection of QDO Problems Derived from HPO of ML Models]{A Collection of Quality Diversity Optimization Problems Derived from Hyperparameter Optimization of Machine Learning Models}

\author{Lennart Schneider}
\affiliation{%
  \institution{LMU Munich}
  \streetaddress{Ludwigstraße 33}
  \city{Munich}
  \country{Germany}
  \postcode{80337}
}
\email{lennart.schneider@stat.uni-muenchen.de}

\author{Florian Pfisterer}
\affiliation{%
  \institution{LMU Munich}
  \streetaddress{Ludwigstraße 33}
  \city{Munich}
  \country{Germany}
  \postcode{80337}
}
\email{florian.pfisterer@stat.uni-muenchen.de}

\author{Janek Thomas}
\affiliation{%
  \institution{LMU Munich}
  \streetaddress{Ludwigstraße 33}
  \city{Munich}
  \country{Germany}
  \postcode{80337}
}
\email{janek.thomas@stat.uni-muenchen.de}

\author{Bernd Bischl}
\affiliation{%
  \institution{LMU Munich}
  \streetaddress{Ludwigstraße 33}
  \city{Munich}
  \country{Germany}
  \postcode{80337}
}
\email{bernd.bischl@stat.uni-muenchen.de}

\renewcommand{\shortauthors}{Schneider et al.}

\begin{abstract}
The goal of Quality Diversity Optimization is to generate a collection of diverse yet high-performing solutions to a given problem at hand.
Typical benchmark problems are, for example, finding a repertoire of robot arm configurations or a collection of game playing strategies.
In this paper, we propose a set of Quality Diversity Optimization problems that tackle hyperparameter optimization of machine learning models - a so far underexplored application of Quality Diversity Optimization.
Our benchmark problems involve novel feature functions, such as interpretability or resource usage of models.
To allow for fast and efficient benchmarking, we build upon YAHPO Gym, a recently proposed open source benchmarking suite for hyperparameter optimization that makes use of high performing surrogate models and returns these surrogate model predictions instead of evaluating the true expensive black box function.
We present results of an initial experimental study comparing different Quality Diversity optimizers on our benchmark problems.
Furthermore, we discuss future directions and challenges of Quality Diversity Optimization in the context of hyperparameter optimization.
\end{abstract}

\ccsdesc[500]{Computing methodologies~Search methodologies}
\keywords{Machine Learning, Hyperparameter Optimization, Quality Diversity Optimization, Benchmarking}

\maketitle

\section{Introduction}

Quality Diversity Optimization (QDO) aims to generate a collection of diverse, high-performing solutions.
Classical algorithms are Novelty Search with Local Competition \cite{lehman2011evolving} and MAP-Elites \cite{mouret2015}, which rely on the concepts of evolutionary computation.
While the development of new QDO algorithms has seen significant progress \cite{fontaine2020,cully2021multi,fontaine2021}, the community lacks larger testbeds of benchmark problems.
In this paper, we propose a set of QDO problems derived from hyperparameter optimization (HPO) of machine learning (ML) models.
So far, HPO has been an underexplored area for the application of QDO.
Our benchmark problems involve feature functions of high practical importance, such as interpretability or resource usage of models, making the application of QDO algorithms particularly relevant due to their illuminating properties.
As an example, consider a random forest \cite{breimann2011} in a binary classification setting.
Depending on the choice of hyperparameters, a random forest can potentially contain many deep trees.
On the one hand, we expect such a random forest to yield good performance, on the other hand, a random forests with thousands of deep trees is not interpretable.
We could for example be interested in relating the performance of a random forest to interpretability measures such as the number of features the model uses or the interaction strength of features.
In general, a random forest that uses few features and has low interaction strength is expected to perform worse than a large and deep forest, but this strongly depends on the concrete data set at hand.
It is therefore of high interest to illuminate the relationship of performance and interpretability and QDO algorithms are very well suited for this task.

\subsection{Hyperparameter Optimization}
An ML \textit{learner} or \textit{inducer} $\inducer$ configured by hyperparameters $\lambdav \in \Lambda$ maps a data set $\D \in \allDatasets$ to a model $\fh$, i.e.,
\begin{equation*}
\inducer : \preimageInducerShort \to \Hspace, 
(\D, \lambdav) \mapsto \fh.
\end{equation*}
HPO methods aim to identify a high-performing hyperparameter configuration (HPC) $\lambdav \in \LamS$ for $\Ilam$ \cite{bischl_hyperparameter_2021}.
The so-called search space $\LamS \subset \Lambda$ is typically a subspace of the set of all possible HPCs: $\LamS = \LamS_1 \times \LamS_2 \times \dots \times \LamS_d,$ where $\LamS_i$ is a bounded subset of the domain of the $i$-th hyperparameter $\Lambda_i$.
$\LamS_i$ can be either continuous, discrete, or categorical. 
It can also dependent on other hyperparameters, meaning that $\LamS_i$ is only active when $\LamS_j$ takes certain values, resulting in a possibly hierarchical search space.
The classical (single-objective) HPO problem is defined as:
\begin{eqnarray}
    \lams \in \argmin_{\lambdav \in \LamS} \GEhlam,
    \label{eq:hpo_objective}
\end{eqnarray}
i.e., the goal is to minimize the estimated generalization error.
This typically involves a costly resampling procedure that can take a significant amount of time (see \cite{bischl_hyperparameter_2021} for further details).
$\GEhlam$ is a black-box function, as it generally has no closed-form mathematical representation, and analytic gradient information is generally not available. 
Therefore, the minimization of $\GEhlam$ forms an \emph{expensive black-box} optimization problem.
Furthermore, it is typically multimodal and noisy, as $\GEhlam$ is only a stochastic estimate of the true unknown generalization error. 

\subsection{Quality Diversity Optimization}
The goal of QDO is to generate a collection of diverse yet high-performing solutions.
Diversity is defined via behavioral niches on so-called feature functions (behavior functions), whereas performance is defined on an objective function (fitness function).
In the following, we assume that a single multi-objective function $f: \Lambda \rightarrow \R^m, m \ge 2$ returns both the objective function as well as the feature functions that span the behavior space.
Without loss of generality, we assume that the first dimension returned by $f$ is the objective function, whereas the dimensions $2, \ldots, m$ are the feature functions.
We denote by $y_{i}$ the i-th output value of $f(\lambdav)$.
We make few assumptions regarding the nature of niches but note that a uniform grid of $k$ niches is classically constructed by dividing the behavior space into equally sized hyperrectangles \cite{mouret2015}.
In this case, a niche $N_{j} \subseteq \R^{m-1}$ is simply defined by lower ($l_{ij}$) and upper ($u_{ij}$) boundaries on each feature function $i \in \{2, \ldots, m\}$, with $l_{ij} = u_{ij-1}$ for $j \in \{2, \ldots, k\}$ .
A point $\lambdav$ ``belongs'' to niche $N_{j}$ if $\forall i \in \{2, \ldots, m\}: y_{i} \in [l_{ij}, u_{ij})$.
The best-performing point of a niche is typically referred to as a so-called elite.
In this paper, we propose several benchmark instances for QDO problems, which we formally introduce below.

\begin{define}[QDO Benchmark Problem]
\label{eq:problem}
A QDO benchmark problem consists of a function $f: \Lambda \rightarrow \R^m, m \ge 2$ returning both the objective function and feature function values, a bounded search space $\tilde{\Lambda}$, and a set of $k$ behavioral niches $\{N_{1}, \ldots N_{k}\}$. $f(\lambdav) = \bm{y}$ returns $m$ values $\bm{y} = (y_{1}, ..., y_{m})^\prime$, where $y_{1}$ is the objective function value and $y_{2}, \ldots, y_{m}$ are feature function values. Based on the niches and their properties (e.g., boundaries), solutions are assigned to these niches and collected in an archive.
\end{define}

\section{Proposed Benchmarks}
We propose a collection of twelve QDO benchmark problems belonging to two different contexts:
First, we are interested in illuminating interpretability (measured by the number of features used in a model and their interaction strength) and performance trade-offs of ML models.
Second, we examine performance trade-offs with respect to model size (memory requirement) and inference time (i.e., time required to make a prediction).
As ML models, we select random forest \cite{breimann2011} and extreme gradient boosting (XGBoost) models \cite{chen2016} in binary classification settings.
As an objective function, we choose the classification error in the case of formulating a minimization problem or the accuracy (1 - classification error) in the case of formulating a maximization problem.
In the following, we assume that the accuracy is to be maximized.
Our benchmark problems rely on YAHPO Gym \cite{pfisterer2022}, a recently proposed open source benchmarking suite for HPO.
YAHPO Gym provides so-called surrogate-based benchmarks: For a given learner and a given data set, a large amount of HPCs (typically sampled uniformly at random from the search space) were evaluated, and surrogate models have been fitted on these data.
When optimizing a benchmark problem (instance) in YAHPO Gym, the costly real black-box evaluation of an HPC is skipped, and instead, the surrogate model's prediction is returned.
This allows for efficient benchmarking of HPO problems, as YAHPO Gym typically requires as little as one millisecond for predicting the performance metric of an HPC and induces minimal memory overhead \cite{pfisterer2022}.
In general, performance estimation of ML models is noisy.
YAHPO Gym can handle both scenarios of providing a deterministic or noisy HPO benchmark.
In the following, we employ YAHPO Gym in the deterministic setting and leave the construction of noisy QDO problems to future research.

We build upon YAHPO Gym's \texttt{iaml\_ranger} and \texttt{iaml\_xgboost} benchmark scenarios.
A scenario is a collection of benchmark instances that share the same learner and search space but contain multiple data sets.
Both \texttt{iaml\_ranger} and \texttt{iaml\_xgboost} allow for HPO on four different data sets with the following OpenML \cite{vanschoren2013openML} IDs: 41146, 40981, 1489, 1067\footnote{These IDs correspond to OpenML data set ids through which data set properties can be queried via \url{https://www.openml.org/d/<id>}.}.

\subsection{Accuracy and Interpretability}
Eight of our proposed QDO benchmark problems belong to the context of performance and interpretability.
We suggest benchmarking all instances of the \texttt{iaml\_ranger} and \texttt{iaml\_xgboost} scenarios because previous benchmarks on YAHPO Gym have shown that these are interesting problems with relevant search spaces.
The search spaces (genotype spaces) are given in \Cref{tab:ss_ranger} and \Cref{tab:ss_xgboost}.
These search spaces are specifically tailored for existing QDO algorithms and implementations, which often do not support categorical parameters or hierarchical dependencies but only numeric parameters.
We discuss this limitation in \Cref{sec:outlook}.
The objective function is given by the accuracy that is to be maximized.
The behavior space is spanned by two feature functions: the number of features (NF) used by a model and their interaction strength (IAS).
These measures were proposed to quantify interpretability for any ML model in a model-agnostic way \cite{molnar2019}.
A feature is regarded as being used by a model if changing the feature changes the prediction of the model.
NF is then estimated via a sampling procedure as described in Algorithm 1 of \cite{molnar2019}.
The IAS is based on accumulated local effects (ALE) \cite{apley2020visualizing} of a model and is given by the scaled approximation error between the ALE main effect model (the sum of first-order ALE effects) and the prediction function of the model.
For more details, see Section 3.2 of \cite{molnar2019}.
A model with low NF and IAS is more interpretable than a model with high NF or high IAS, but optimal performance is often reflected in either a high NF or IAS or both.
However, this strongly depends on the data set at hand; as ML models can always overfit, resulting in bad generalization performance, simply opting for models having a high NF and IAS is prone to poor performance.
It is therefore a priori unknown whether low or high NF / IAS will result in good performance, making the application of QDO algorithms particularly attractive due to the illuminating aspect of QDO.
In line with most QDO algorithms and their implementations, we propose to define niches, using a discrete uniform grid archive with ranges $[0, p]$ for NF and $[0, 1]$ for IAS.
Here, $p$ is the number of features present in a data set (e.g., $14$ for data set 40981).
We propose to use a bin size of $p + 1$ for NF (i.e., models in each bin use exactly $0, 1, \dots$ or $p$ features) and $100$ for IAS, resulting in $2100$, $1500$, $600$, and $2200$ niches overall for the data sets above.
We summarize all benchmark problems in \Cref{tab:problems}.
In \Cref{fig:heatmaps}, we visualize the exemplary \texttt{iaml\_ranger\_40981} problem via a heatmap of the accuracy (color) of different elites, which allows for visually inspecting the alignment of the feature functions and objective function.
We present similar heatmaps for all remaining benchmark problems in \Cref{sec:appendix}.

\subsection{Accuracy and Resource Usage}
\begin{figure}[h]
\minipage{0.23\textwidth}
  \includegraphics[width=\linewidth]{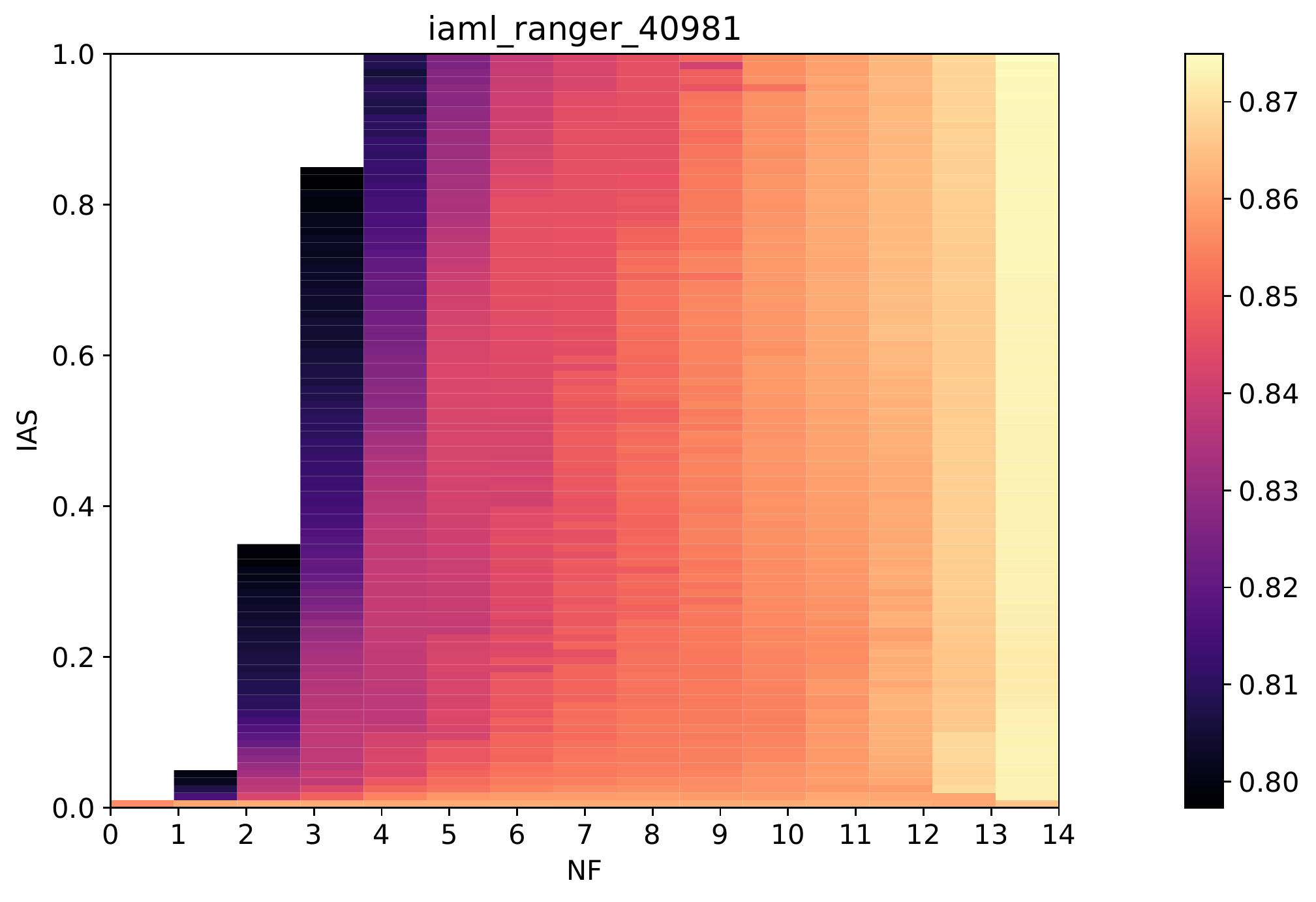}
\endminipage\hfill
\minipage{0.23\textwidth}%
  \includegraphics[width=\linewidth]{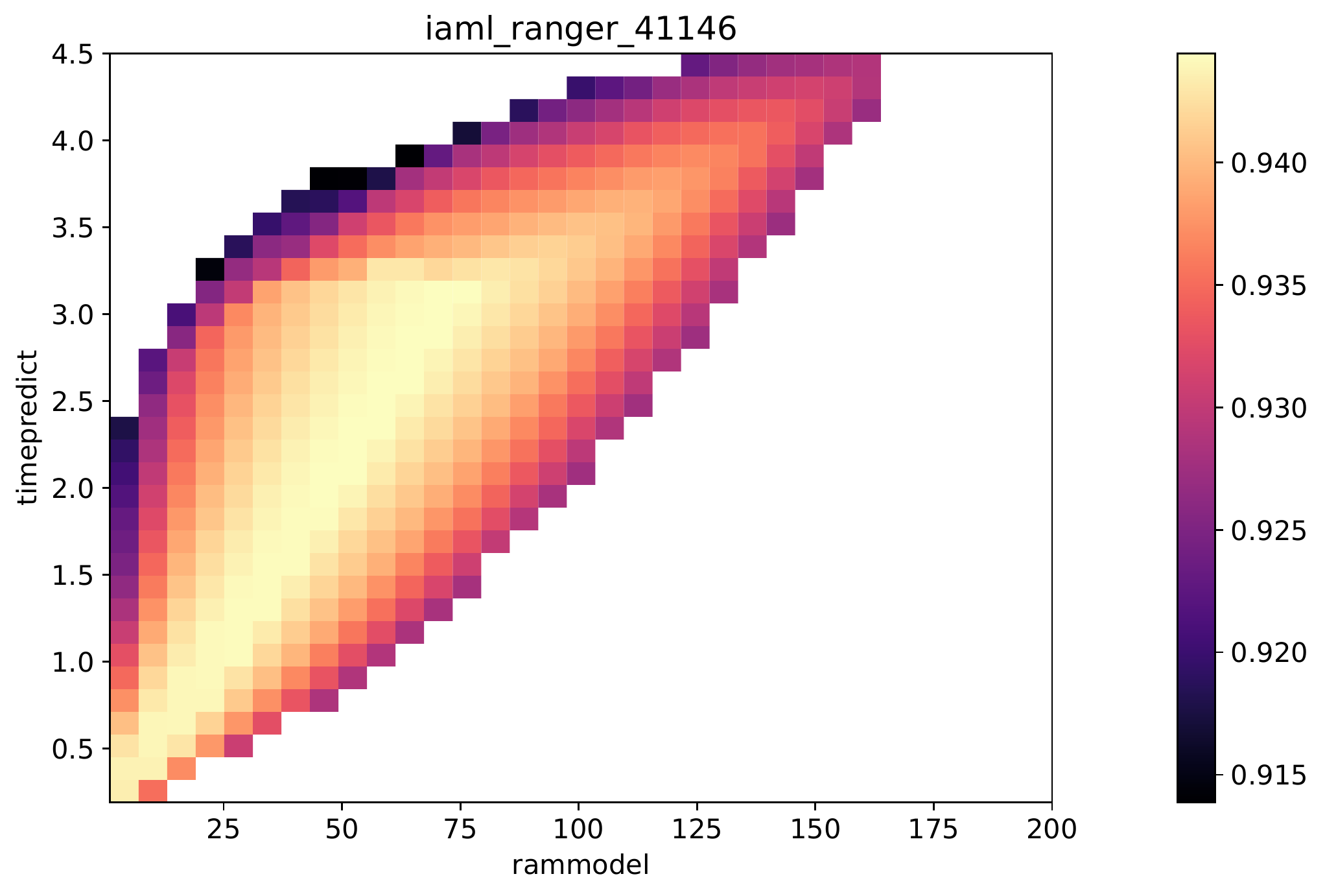}
\endminipage
\caption{Heatmaps of the \texttt{iaml\_ranger\_40981} interpretability (left) and \texttt{iaml\_ranger\_41146} resource usage (right) benchmark problems. The color gradient represents the accuracy (objective function), with black indicating lowest accuracy and bright yellow/white indicating highest accuracy.}\label{fig:heatmaps}
\end{figure}

Four of our proposed QDO problems belong to the context of performance and resource usage.
We suggest benchmarking all instances of the \texttt{iaml\_ranger} scenario.
The search space is again given in \Cref{tab:ss_ranger}.
The objective function is given by the accuracy that is to be maximized.
The behavior space is spanned by two feature functions: the size of the model if stored on disk (in MB) and the time required to make a prediction on the same data set the model has been trained on (in seconds).
These measures are typically of interest in the context of deploying ML models in production environments on different hardware with varying restrictions regarding RAM and latency of predictions \cite{nan2016pruning}.
A larger random forest typically consists of plenty and deep trees, and prediction time is both affected by the depth of the individual trees and the number of trees in the forest.
In general, a large random forest with deep trees should result in a better performance, but this again depends on the data set at hand.
As before, we propose to define niches via a discrete uniform grid archive (for more details, see \Cref{tab:problems}).
In \Cref{fig:heatmaps}, we visualize the exemplary \texttt{iaml\_ranger\_41146} problem via a heatmap of the accuracy (color) of different elites, which allows for visually inspecting the alignment of the feature functions and objective function.
We present similar heatmaps for all remaining benchmark problems in \Cref{sec:appendix}.

\begin{table}[h]
  \centering
  \caption{Search space of \texttt{iaml\_ranger} scenarios.}
  \label{tab:ss_ranger}
  \small
  \begin{threeparttable}
  \begin{tabular}{lrrr}
  \toprule
  \textbf{Hyperparameter} & \textbf{Type} & \textbf{Range} & \textbf{Trafo}\\
  \midrule
  num.trees & integer & [1, 2000] & \\
  mtry.ratio & continuous & [0, 1] & \\
  min.node.size & integer & [1, 100] & \\
  sample.fraction & continuous & [0.1, 1] & \\
  \bottomrule
  \end{tabular}
  \end{threeparttable}
\end{table}

\begin{table}[h]
  \centering
  \caption{Search space of \texttt{iaml\_xgboost} scenarios.}
  \label{tab:ss_xgboost}
  \small
  \begin{threeparttable}
  \begin{tabular}{lrrr}
  \toprule
  \textbf{Hyperparameter} & \textbf{Type} & \textbf{Range} & \textbf{Trafo} \\
  \midrule
  alpha & continuous & [1e-04, 1000] & log \\
  lambda & continuous & [1e-04, 1000] & log\\ 
  nrounds & integer & [3, 2000] & log \\ 
  subsample & continuous & [0.1, 1] & \\ 
  colsample\_bylevel & continuous & [0.01, 1] & \\ 
  colsample\_bytree & continuous & [0.01, 1] & \\ 
  eta & continuous & [1e-04, 1] & log \\ 
  gamma & continuous & [1e-04, 7] & log \\ 
  max\_depth & integer & [1, 15] & \\ 
  min\_child\_weight & continuous & [$\exp(1)$, 150] & log \\ 
  \bottomrule
  \end{tabular}
  \begin{tablenotes}
  \item "log" in the Trafo column indicates that this parameter is optimized on a logarithmic scale, i.e., the range is given by $[\log(\mathrm{lower}), \log(\mathrm{upper})]$, and values are re-transformed via the exponential function prior to evaluating the HPC.
  \end{tablenotes}
  \end{threeparttable}
\end{table}

\begin{table*}[h]
  \centering
  \caption{Summary of our QDO benchmark problems.}
  \label{tab:problems}
  \small
  \begin{threeparttable}
  \begin{tabular}{lrrrrrr}
  \toprule
  \textbf{Problem} & \textbf{Learner} & \textbf{Data set} & \textbf{Search space} & \textbf{Objective} & \textbf{Features} & \textbf{Niches}\\
  \midrule
  Illuminating interpretability & & & & & & \\
  \midrule
  \texttt{iaml\_ranger\_41146}  & rf            & 41146  & \Cref{tab:ss_ranger}  & acc & NF + IAS & uniform on $[0, 20] \times [0, 1]$ $21, 100$ bins\\
  \texttt{iaml\_ranger\_40981}  & rf            & 40981  & \Cref{tab:ss_ranger}  & acc & NF + IAS & uniform on $[0, 14] \times [0, 1]$ $15, 100$ bins\\
  \texttt{iaml\_ranger\_1489}   & rf            & 1489   & \Cref{tab:ss_ranger}  & acc & NF + IAS & uniform on $[0, 5]  \times [0, 1]$ $6, 100$ bins\\
  \texttt{iaml\_ranger\_1067}   & rf            & 1067   & \Cref{tab:ss_ranger}  & acc & NF + IAS & uniform on $[0, 21] \times [0, 1]$ $22, 100$ bins\\
  \texttt{iaml\_xgboost\_41146} & XGBoost       & 41146  & \Cref{tab:ss_xgboost} & acc & NF + IAS & uniform on $[0, 20] \times [0, 1]$ $21, 100$ bins\\
  \texttt{iaml\_xgboost\_40981} & XGBoost       & 40981  & \Cref{tab:ss_xgboost} & acc & NF + IAS & uniform on $[0, 14] \times [0, 1]$ $15, 100$ bins\\
  \texttt{iaml\_xgboost\_1489}  & XGBoost       & 1489   & \Cref{tab:ss_xgboost} & acc & NF + IAS & uniform on $[0, 5]  \times [0, 1]$ $6, 100$ bins\\
  \texttt{iaml\_xgboost\_1067}  & XGBoost       & 1067   & \Cref{tab:ss_xgboost} & acc & NF + IAS & uniform on $[0, 21] \times [0, 1]$ $22, 100$ bins\\
  \midrule
  Illuminating resource usage & & & & & & \\
  \midrule
  \texttt{iaml\_ranger\_41146}  & rf            & 41146  & \Cref{tab:ss_ranger}  & acc & rm + tp & uniform on $[1, 200] \times [0.19, 4.5]$ $33, 33$ bins\\
  \texttt{iaml\_ranger\_40981}  & rf            & 40981  & \Cref{tab:ss_ranger}  & acc & rm + tp & uniform on $[1, 40] \times [0.10, 0.65]$ $33, 33$ bins\\
  \texttt{iaml\_ranger\_1489}   & rf            & 1489   & \Cref{tab:ss_ranger}  & acc & rm + tp & uniform on $[1, 200] \times [0.19, 4.5]$ $33, 33$ bins\\
  \texttt{iaml\_ranger\_1067}   & rf            & 1067   & \Cref{tab:ss_ranger}  & acc & rm + tp & uniform on $[1, 78] \times [0.13, 1.55]$ $33, 33$ bins\\
  \bottomrule
  \end{tabular}
  \begin{tablenotes}
  \item "rf" = random forest. "acc" = accuracy. "rm" = rammodel. "tp" = timepredict. Niches of the uniform grid archive are constructed by dividing each dimension of the feature space as indicated in the Niches column into equally spaced bins, forming hyperrectangles.
  \end{tablenotes}
  \end{threeparttable}
\end{table*}

\subsection{Benchmark Experiments and Results}
\begin{figure}[h]
\includegraphics[width=\linewidth]{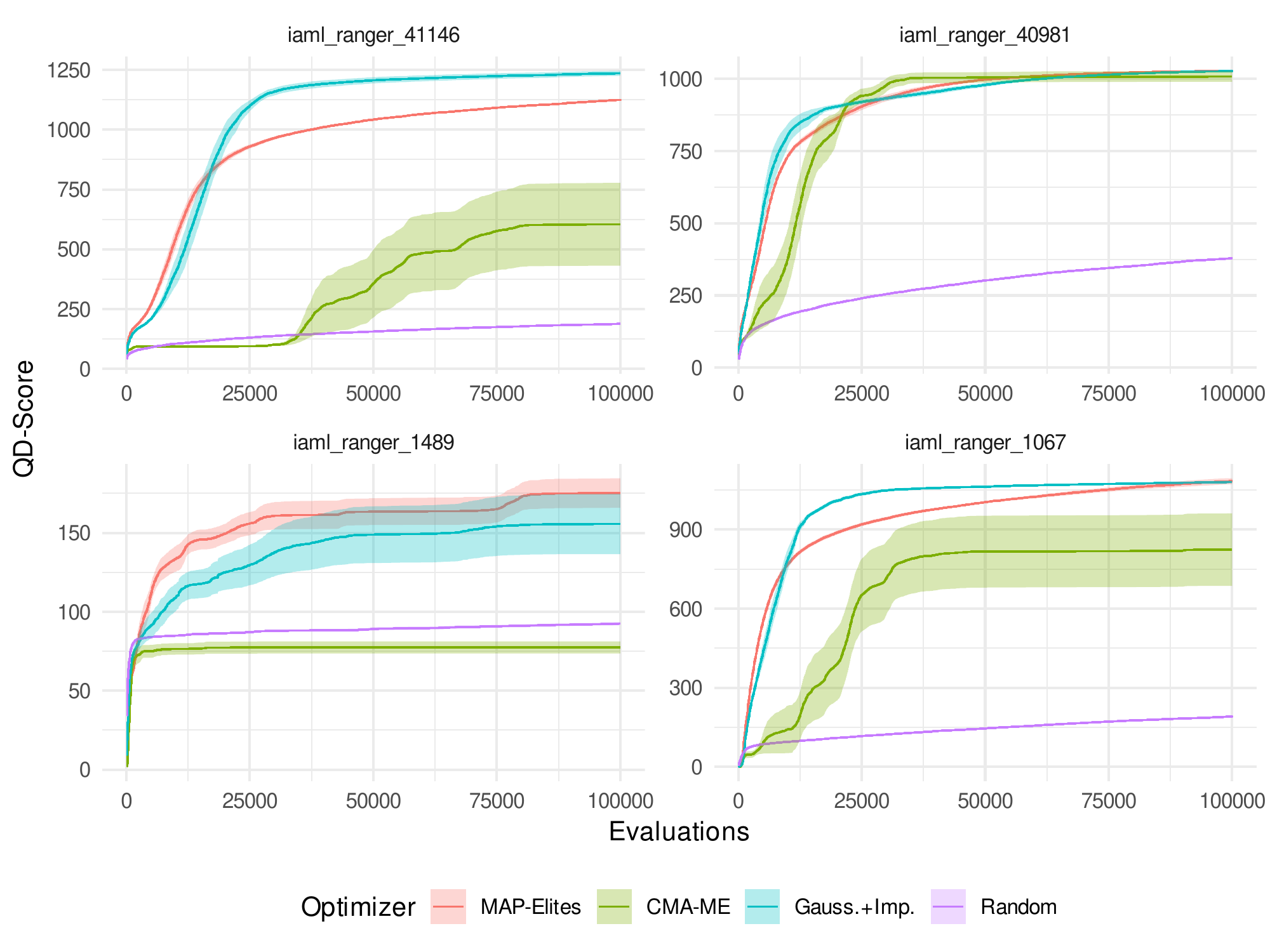}
\caption{Anytime QD-Score for the \texttt{iaml\_ranger} benchmarks (interpretability). Averaged over 10 replications. Ribbons represent standard errors.}\label{fig:results_iaml_ranger_inter}
\end{figure}

\begin{figure}[h]
\includegraphics[width=\linewidth]{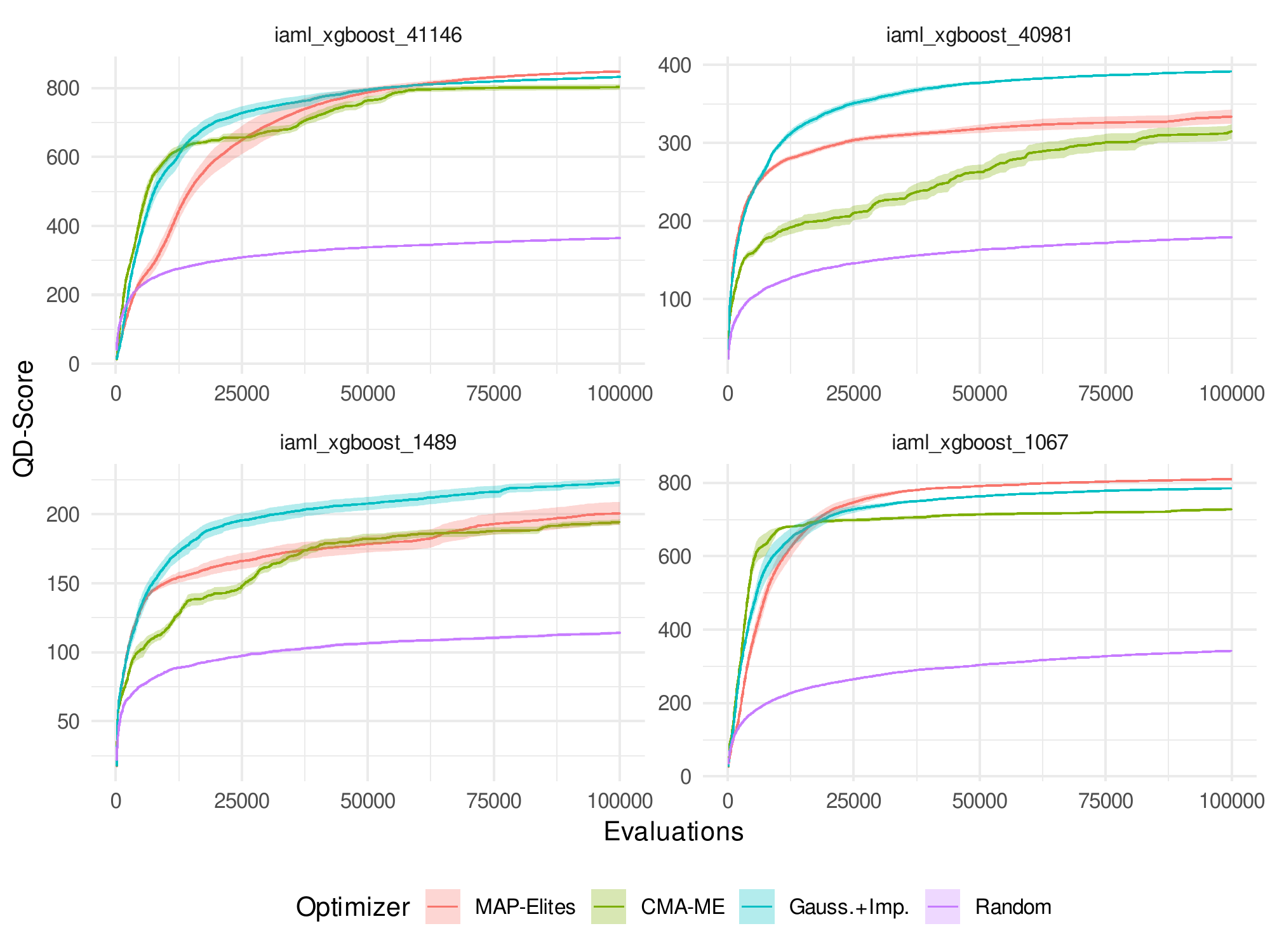}
\caption{Anytime QD-Score for the \texttt{iaml\_xgboost} benchmarks (interpretability). Averaged over 10 replications. Ribbons represent standard errors.}\label{fig:results_iaml_xgboost_inter}
\end{figure}

\begin{figure}[h]
\includegraphics[width=\linewidth]{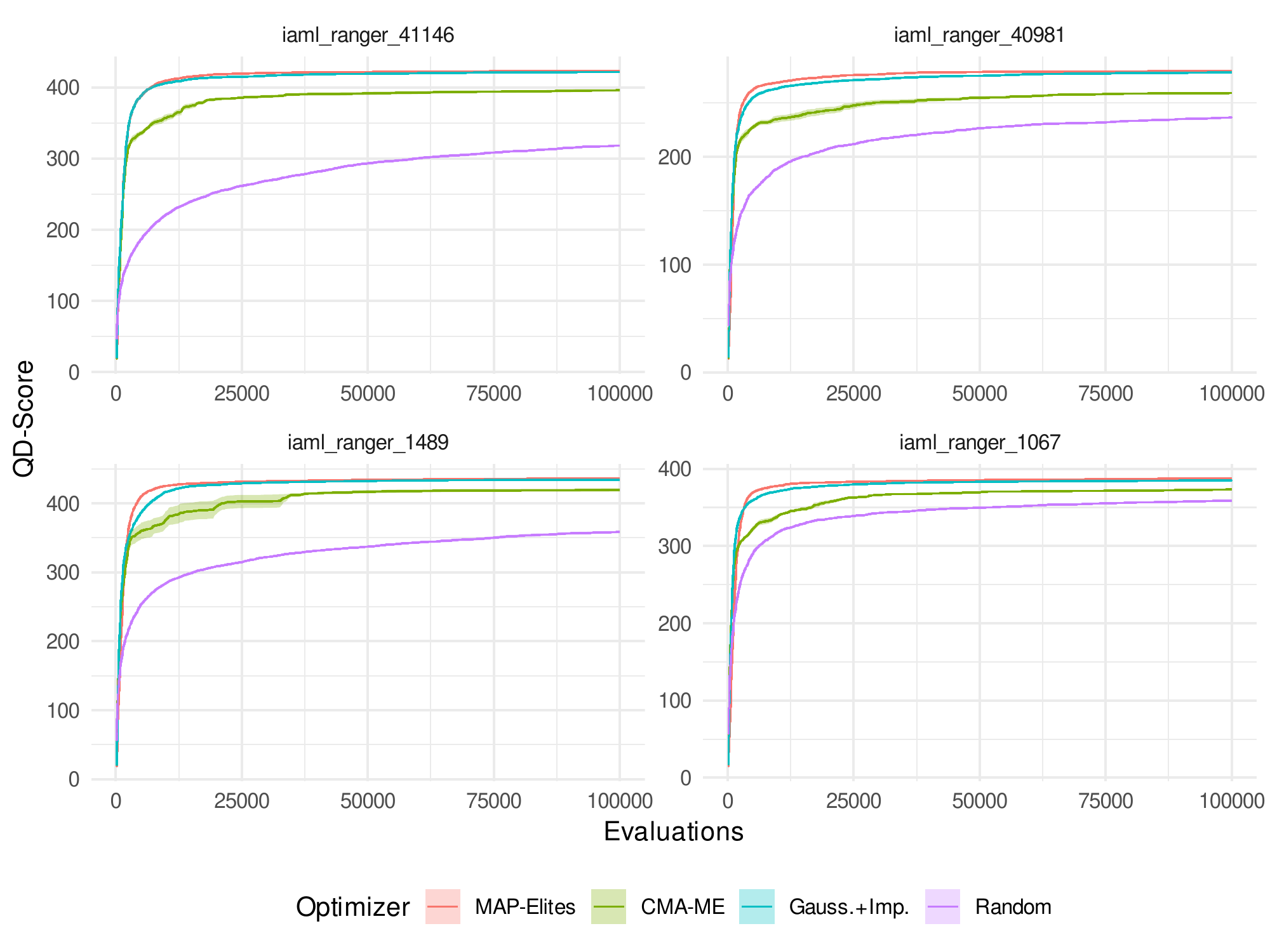}
\caption{Anytime QD-Score for the \texttt{iaml\_ranger} benchmarks (resource usage). Averaged over 10 replications. Ribbons represent standard errors.}\label{fig:results_iaml_ranger_hardware}
\end{figure}

We benchmark a basic variant of MAP-Elites \cite{mouret2015}, a basic variant of CMA-ME \cite{fontaine2020}, and Random Search (i.e., sampling HPCs uniformly at random).
To illustrate the potential of fast prototyping of new methods, we also include an optimizer that is a mix between MAP-Elites and CMA-ME, where half of the evaluations of a batch are proposed using a Gaussian emitter and the other half using an improvement emitter, which we abbreviate as ``Gauss.+Imp.''.
We implement all optimizers in pyribs \cite{pyribs}.
All optimizer use a batch size of $100$ and run for $1000$ iterations, resulting in $100000$ evaluations in total.
For MAP-Elites, we use $\sigma_{0} = 0.1$ as the standard deviation of the Gaussian distribution. For CMA-ME, we set $\sigma_{0} = 0.1$ as the initial step size and use ``filter`` as a selection rule.
All optimizers use a single emitter.
We want to note that we did not perform any tuning of hyperparameters of the optimizers themselves.
Therefore, results are preliminary and better performance of an optimizer could be obtained by more carefully chosen design choices.
All runs are replicated ten times with different random seeds.
We always normalize all parameters of the search space to the unit cube and optimize within these normalized bounds (during evaluation of an HPC, parameters are then re-transformed to their original scale).
Note that integer-valued parameters are treated as continuous. However, prior to evaluating an HPC, those parameter values are then rounded to the nearest integer.
As performance metrics, we report the coverage (percent of niches occupied with a solution), the QD-Score \cite{pugh2016} (sum of best accuracies of occupied niches), and the overall best accuracy found.
Results for the interpretability context are given in \Cref{tab:results_iaml_ranger_inter} for the \texttt{iaml\_ranger} scenarios and in \Cref{tab:results_iaml_xgboost_inter} for the \texttt{iaml\_xgboost} scenarios.
Results for the resource usage context are given in \Cref{tab:results_iaml_ranger_hardware}.
We also visualize the anytime QD-Score in \Cref{fig:results_iaml_ranger_inter}, \Cref{fig:results_iaml_xgboost_inter}, and \Cref{fig:results_iaml_ranger_hardware}.
In general, MAP-Elites shows the strongest performance but is sometimes outperformed by the Gauss.+Imp. optimizer.
The performance of CMA-ME varies strongly between benchmark problems, whereas MAP-Elites and Gauss.+Imp. appear to be more consistent in their performance.
Random Search performs overall poorly, indicating that structural information of the problems can be efficiently leveraged by more sophisticated optimizers.
We release all our code for running the benchmarks and analyzing results via the following GitHub repository: \url{https://github.com/slds-lmu/qdo_yahpo}.

\begin{table*}[h]
  \centering
  \caption{Benchmark results on \texttt{iaml\_ranger} (interpretability).}
  \label{tab:results_iaml_ranger_inter}
  \small
  \begin{threeparttable}
  \begin{tabular}{|l|rrr|rrr|}
                        & data set: 41146          &                   &                       & data set: 40981      &                   &   \\
  \midrule
  \textbf{Algorithm}    & \textbf{Coverage \%}     & \textbf{QD-Score} & \textbf{Max Acc}      & \textbf{Coverage \%} & \textbf{QD-Score} & \textbf{Max Acc}\\
  \midrule
  MAP-Elites            & 58.95                    & 1124.69           & \textbf{0.9444}       & \textbf{80.91}       & \textbf{1026.84}  & \textbf{0.8747}  \\
  CMA-ME                & 31.43                    & 604.66            & 0.9444                & 79.03                & 1007.62           & 0.8745  \\
  Gauss.+Imp.           & \textbf{64.61}           & \textbf{1235.72}  & \textbf{0.9444}       & 80.63                & 1026.35           & 0.8746  \\
  Random                & 9.69                     & 188.52            & 0.9443                & 29.39                & 378.26            & 0.8746  \\
  \midrule
                        & data set: 1489           &                   &                       & data set: 1067       &                   &   \\
  \midrule
  \textbf{Algorithm}    & \textbf{Coverage \%}     & \textbf{QD-Score} & \textbf{Max Acc}      & \textbf{Coverage \%} & \textbf{QD-Score} & \textbf{Max Acc}\\
  \midrule
  MAP-Elites            & \textbf{34.73}           & \textbf{175.24}   & \textbf{0.9112}       & \textbf{58.10}       & \textbf{1083.08}  & \textbf{0.8706}  \\
  CMA-ME                & 14.57                    & 77.41             & 0.9109                & 43.87                & 823.86            & 0.8700  \\
  Gauss.+Imp.           & 30.83                    & 155.62            & 0.9111                & 57.57                & 1080.00           & 0.8705  \\
  Random                & 17.63                    & 92.51             & 0.9109                & 10.22                & 191.19            & 0.8705  \\
  \bottomrule
  \end{tabular}
  \begin{tablenotes}
  \item Averaged over 10 replications. "Max Acc" = maximum accuracy. Best performance is highlighted in bold.
  \end{tablenotes}
  \end{threeparttable}
\end{table*}

\begin{table*}[h]
  \centering
  \caption{Benchmark results on \texttt{iaml\_xgboost} (interpretability).}
  \label{tab:results_iaml_xgboost_inter}
  \small
  \begin{threeparttable}
  \begin{tabular}{|l|rrr|rrr|}
                        & data set: 41146          &                   &                       & data set: 40981      &                   &   \\
  \midrule
  \textbf{Algorithm}    & \textbf{Coverage \%}     & \textbf{QD-Score} & \textbf{Max Acc}      & \textbf{Coverage \%} & \textbf{QD-Score} & \textbf{Max Acc}\\
  \midrule
  MAP-Elites            & \textbf{47.94}           & \textbf{847.67}   & \textbf{0.9446}       & 26.00                & 333.70            & \textbf{0.8870}  \\
  CMA-ME                & 45.38                    & 802.77            & 0.9444                & 24.61                & 314.66            & 0.8849  \\
  Gauss.+Imp.           & 47.03                    & 831.91            & 0.9446                & \textbf{30.76}       & \textbf{391.40}   & 0.8868  \\
  Random                & 20.10                    & 364.31            & 0.9432                & 14.07                & 179.34            & 0.8822  \\
  \midrule
                        & data set: 1489           &                   &                       & data set: 1067       &                   &   \\
  \midrule
  \textbf{Algorithm}    & \textbf{Coverage \%}     & \textbf{QD-Score} & \textbf{Max Acc}      & \textbf{Coverage \%} & \textbf{QD-Score} & \textbf{Max Acc}\\
  \midrule
  MAP-Elites            & 38.85                    & 200.69            & \textbf{0.9147}       & \textbf{42.90}       & \textbf{810.19}   & \textbf{0.8824}  \\
  CMA-ME                & 37.92                    & 194.44            & 0.9132                & 38.47                & 728.03            & 0.8814  \\
  Gauss.+Imp.           & \textbf{43.52}           & \textbf{223.11}   & 0.9147                & 41.55                & 785.12            & 0.8821  \\
  Random                & 22.55                    & 114.04            & 0.9052                & 18.19                & 342.40            & 0.8736  \\
  \bottomrule
  \end{tabular}
  \begin{tablenotes}
  \item Averaged over 10 replications. "Max Acc" = maximum accuracy. Best performance is highlighted in bold.
  \end{tablenotes}
  \end{threeparttable}
\end{table*}

\begin{table*}[h]
  \centering
  \caption{Benchmark results on \texttt{iaml\_ranger} (resource usage).}
  \label{tab:results_iaml_ranger_hardware}
  \small
  \begin{threeparttable}
  \begin{tabular}{|l|rrr|rrr|}
                        & data set: 41146          &                   &                       & data set: 40981      &                   &   \\
  \midrule
  \textbf{Algorithm}    & \textbf{Coverage \%}     & \textbf{QD-Score} & \textbf{Max Acc}      & \textbf{Coverage \%} & \textbf{QD-Score} & \textbf{Max Acc}\\
  \midrule
  MAP-Elites            & \textbf{41.50}           & \textbf{422.99}   & \textbf{0.9444}       & \textbf{29.44}       & \textbf{279.65}   & 0.8749  \\
  CMA-ME                & 38.82                    & 395.85            & 0.9444                & 27.29                & 259.23            & 0.8749  \\
  Gauss.+Imp.           & 41.36                    & 421.55            & 0.9444                & 29.30                & 278.35            & \textbf{0.8749}  \\
  Random                & 31.25                    & 318.20            & 0.9443                & 24.92                & 236.56            & 0.8746  \\
  \midrule
                        & data set: 1489           &                   &                       & data set: 1067       &                   &   \\
  \midrule
  \textbf{Algorithm}    & \textbf{Coverage \%}     & \textbf{QD-Score} & \textbf{Max Acc}      & \textbf{Coverage \%} & \textbf{QD-Score} & \textbf{Max Acc}\\
  \midrule
  MAP-Elites            & \textbf{44.52}           & \textbf{435.75}   & \textbf{0.9112}       & \textbf{41.04}       & \textbf{387.44}   & \textbf{0.8709}  \\
  CMA-ME                & 42.84                    & 419.21            & 0.9112                & 39.53                & 373.20            & 0.8709  \\
  Gauss.+Imp.           & 44.34                    & 434.06            & 0.9112                & 40.77                & 384.92            & 0.8709  \\
  Random                & 36.74                    & 358.33            & 0.9109                & 38.06                & 358.90            & 0.8705  \\
  \bottomrule
  \end{tabular}
  \begin{tablenotes}
  \item Averaged over 10 replications. "Max Acc" = maximum accuracy. Best performance is highlighted in bold.
  \end{tablenotes}
  \end{threeparttable}
\end{table*}

\section{Outlook}\label{sec:outlook}

The benchmark problems proposed in this paper pose novel applications for QDO methods. 
QDO is interesting for HPO problems in contexts where users are interested in criteria that go beyond traditional performance metrics.
This can be highly relevant in the context of ML models that should be interpretable to a certain degree or ML models that must be deployed across a diverse set of systems, ranging from FPGAs to compute clusters.
We propose several such benchmarks but hope that HPO problems for QDO can go far beyond what is proposed in the context of this work.
Furthermore, we identify two properties of existing HPO algorithms and implementation that might help to drive greater adaptation of QDO methods. 

\subsection{Mixed Search Spaces}
Available implementations of QDO methods, such as pyribs, consider only continuous search spaces.
In many practical applications, search spaces include categorical and conditionally active hyperparameters -- so-called hierarchical, mixed search spaces \cite{thornton2013auto}.
While existing methods can theoretically be extended to such spaces, a lack of available implementations might inhibit adoption of QDO methods.
Simultaneously, this might also open up a richer set of readily available problem instances, and our proposed benchmarks can be trivially extended to include such search spaces \footnote{for the full search spaces of our benchmark problems, see \cite{pfisterer2022}}.

\subsection{Overlapping Niches}
Similarly, while not formally required, niches in the QDO literature have traditionally been defined as being disjoint, i.e., a point can only belong to a single niche.
Real-world applications in the context of HPO provide an interesting field of study for methods that go beyond those assumptions.
As an example, consider again the setting where an ML model should be deployed across diverse hardware.
In this context, niches could overlap, as a smaller model (e.g., fitting on an FPGA or microcontroller) is always also a candidate for larger devices (e.g., fitting on a laptop or workstation).

The field of hardware-aware neural architecture search \cite{benmeziane21} seeks to find neural network architectures that are optimal for specific hardware.
If such models are now required for a variety of different hardware architectures, QDO could be applied to find models for each of those niches while simultaneously exploiting similarities between architectures.
In this setting, niches cannot be defined on simple feature functions, as a solution can fall into multiple niches that defy definition based on simple rules. 

Such applications form an interesting new playground for QDO methods, especially when considering that HPO problems are usually considered to be \emph{expensive}.
This might provide a valuable new avenue for future method development towards approaches that require fewer function evaluations \cite{kent20} or that can efficiently make use of evaluations at multiple fidelity levels \cite{li17}.

\begin{acks}
The authors of this work take full responsibilities for its content.
Lennart Schneider is supported by the Bavarian Ministry of Economic Affairs, Regional Development and Energy through the Center for Analytics - Data – Applications (ADACenter) within the framework of BAYERN DIGITAL II (20-3410-2-9-8).
This work was supported by the German Federal Ministry of Education and Research (BMBF) under Grant No. 01IS18036A.
\end{acks}

\bibliographystyle{ACM-Reference-Format}
\bibliography{refs}

\appendix
\section{Heatmaps for all Benchmark Problems}\label{sec:appendix}
\begin{figure}[h]
\minipage{0.23\textwidth}
  \includegraphics[width=\linewidth]{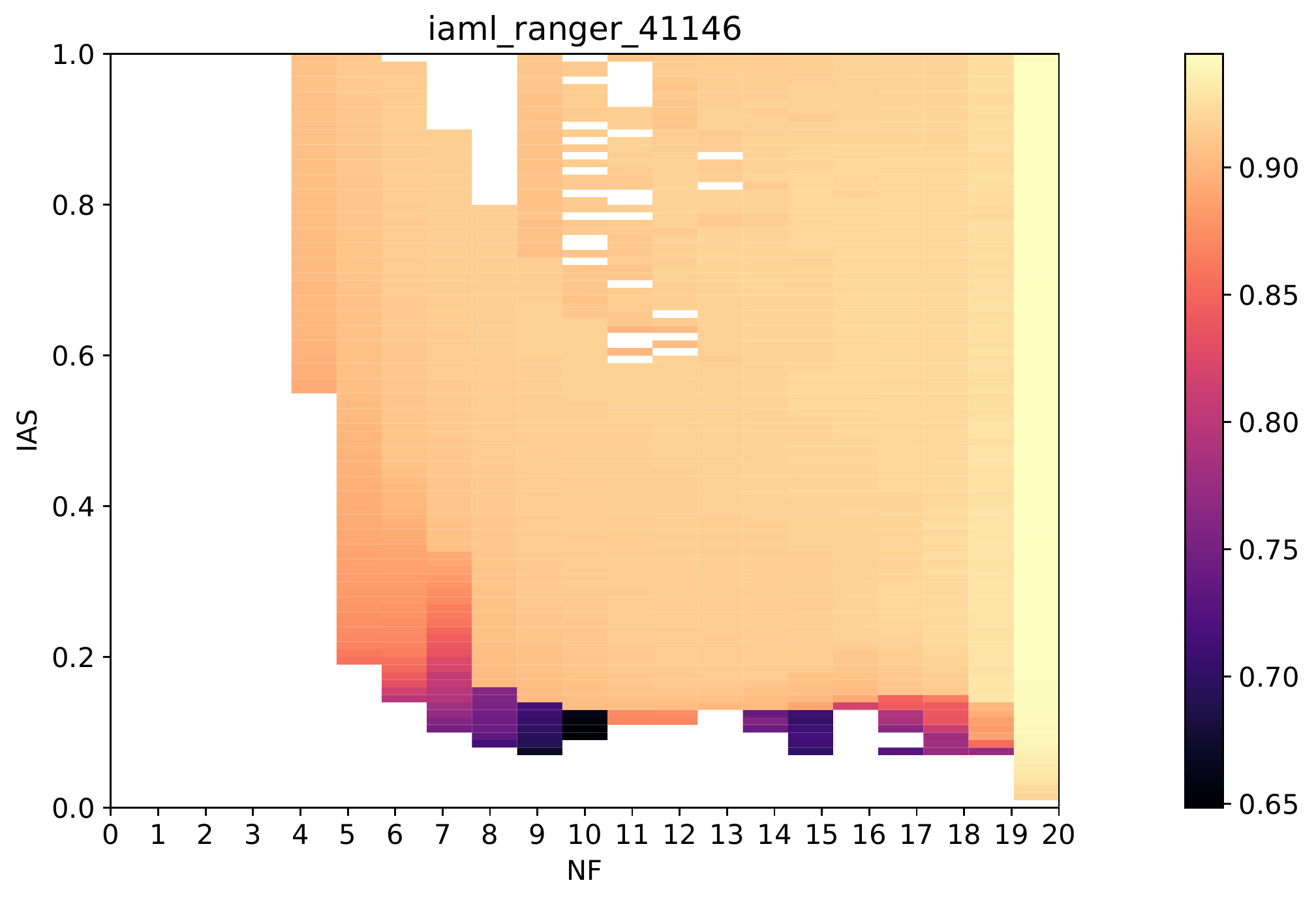}
\endminipage\hfill
\minipage{0.23\textwidth}%
  \includegraphics[width=\linewidth]{Plots/iaml_ranger_ias_nf_40981.pdf}
\endminipage\\
\minipage{0.23\textwidth}
  \includegraphics[width=\linewidth]{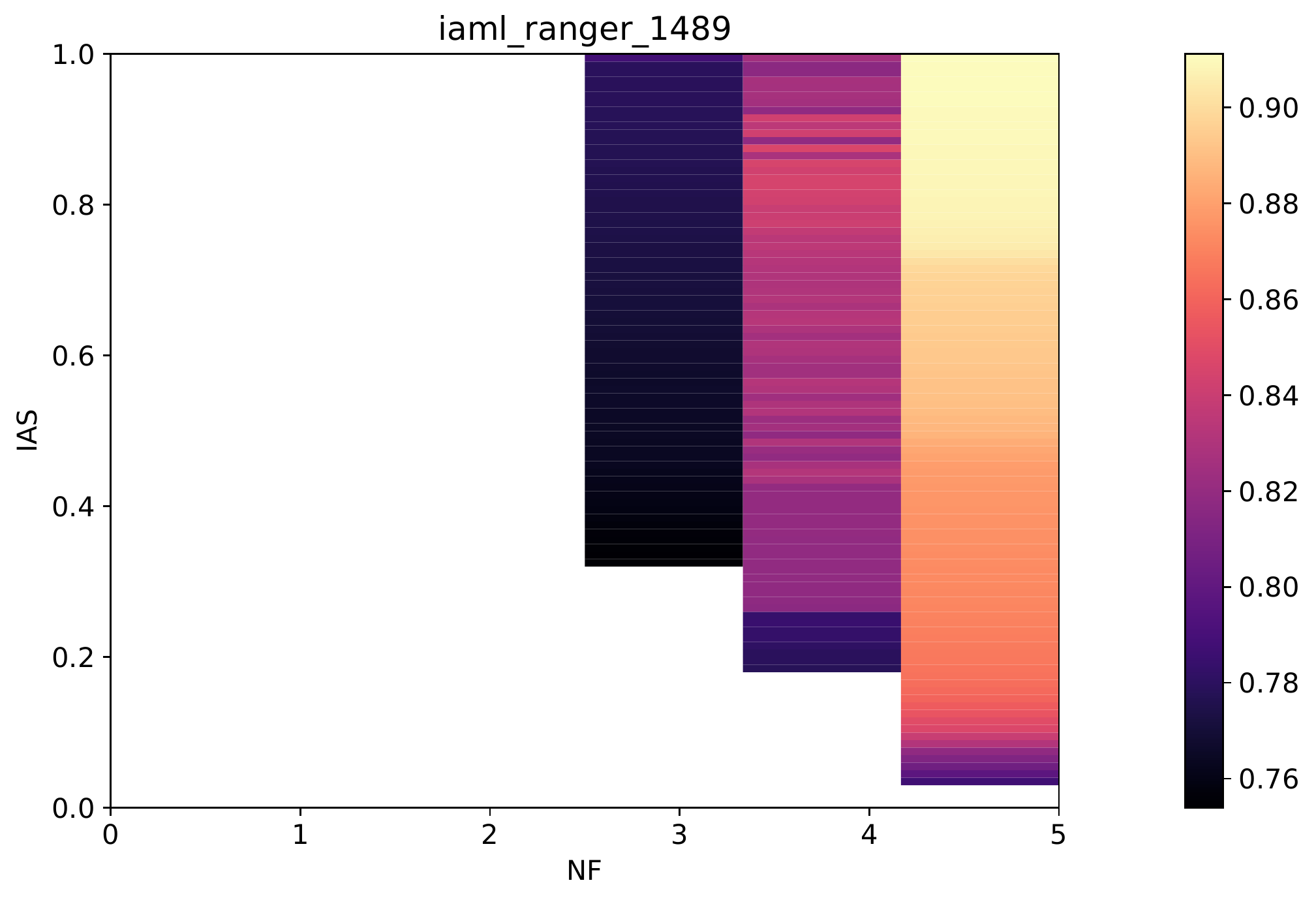}
\endminipage\hfill
\minipage{0.23\textwidth}%
  \includegraphics[width=\linewidth]{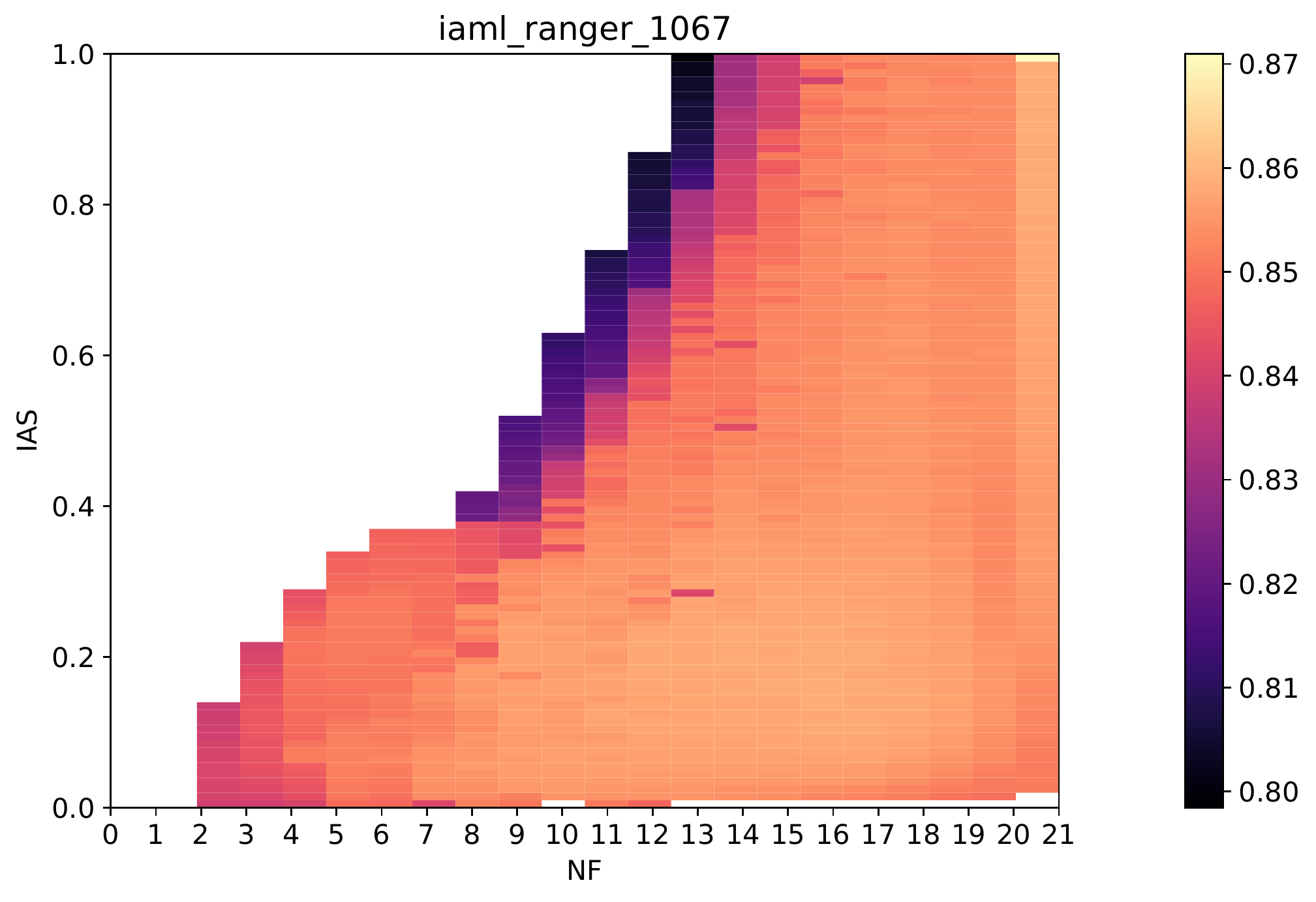}
\endminipage
\caption{Heatmaps of the \texttt{iaml\_ranger} interpretability benchmark problems.}\label{fig:heatmaps_ranger_interpretability}
\end{figure}

\begin{figure}[h]
\minipage{0.23\textwidth}
  \includegraphics[width=\linewidth]{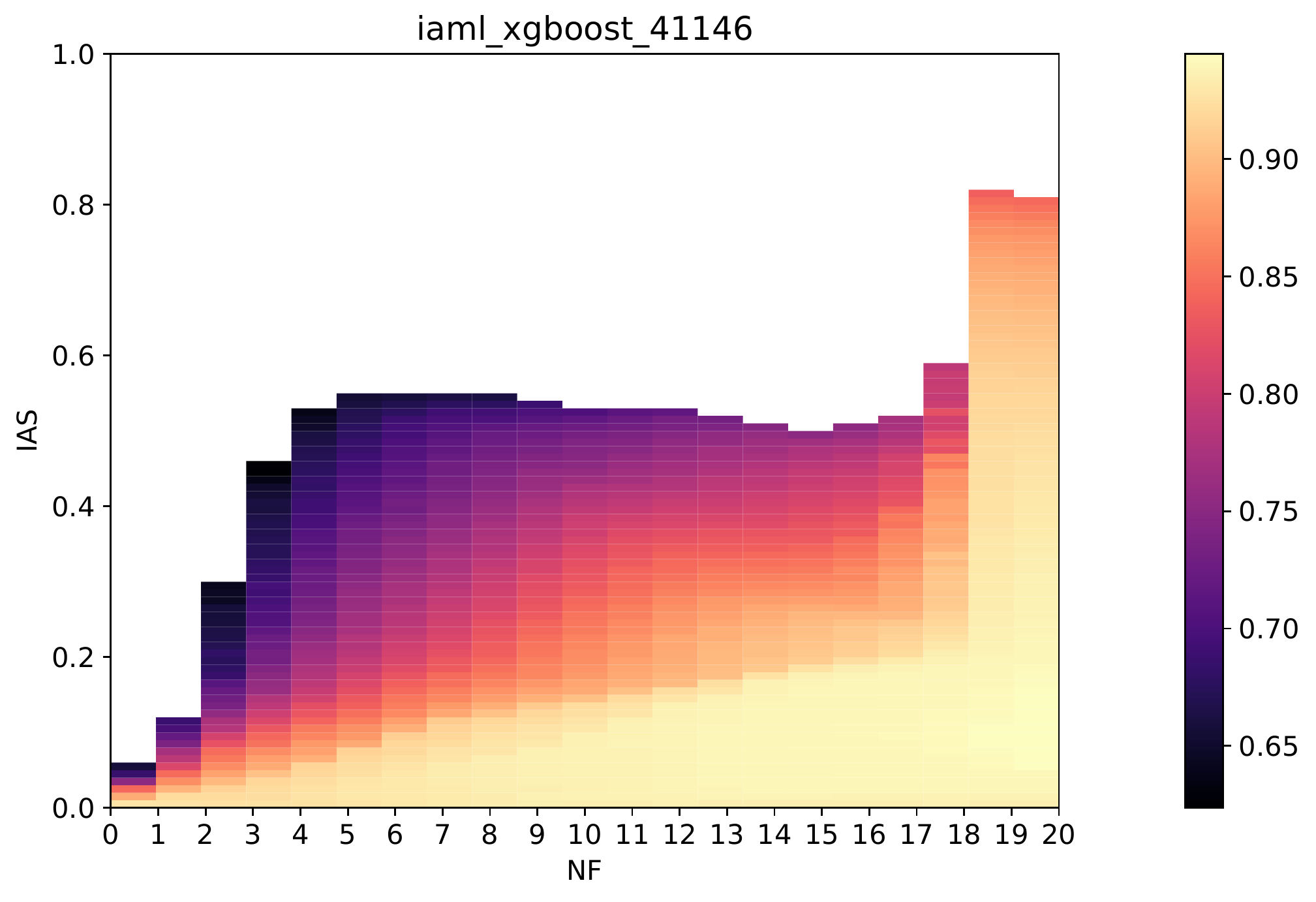}
\endminipage\hfill
\minipage{0.23\textwidth}%
  \includegraphics[width=\linewidth]{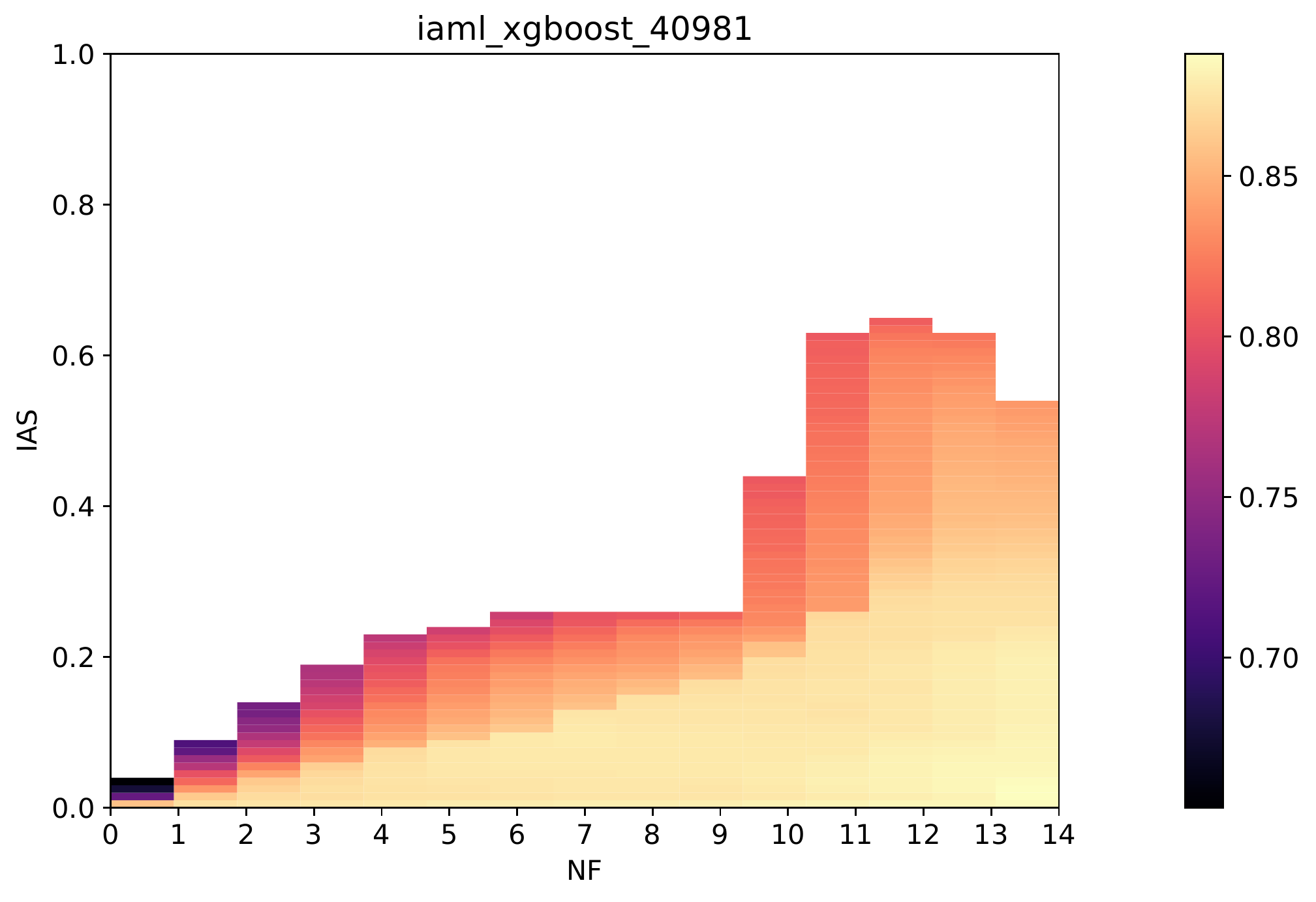}
\endminipage\\
\minipage{0.23\textwidth}
  \includegraphics[width=\linewidth]{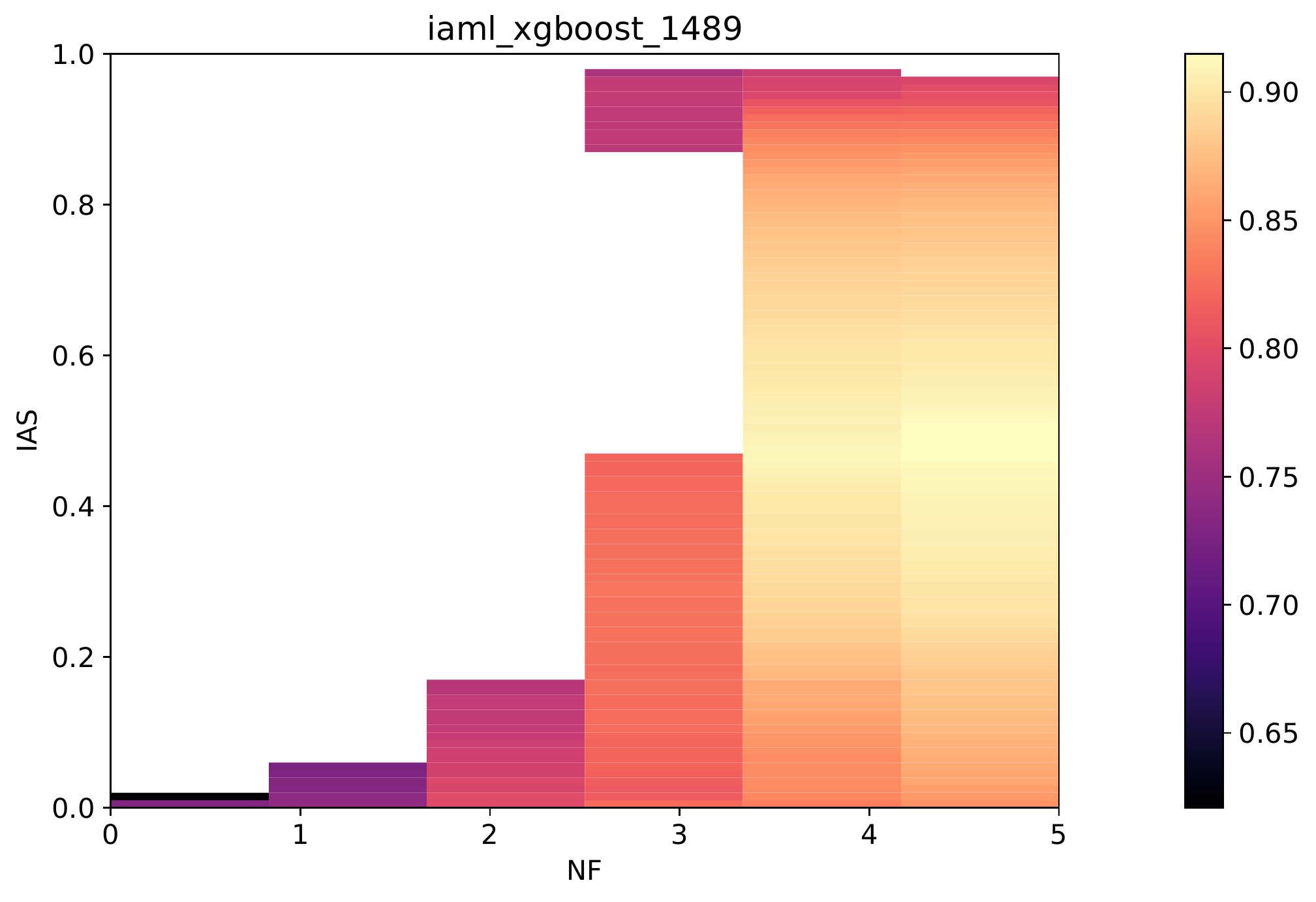}
\endminipage\hfill
\minipage{0.23\textwidth}%
  \includegraphics[width=\linewidth]{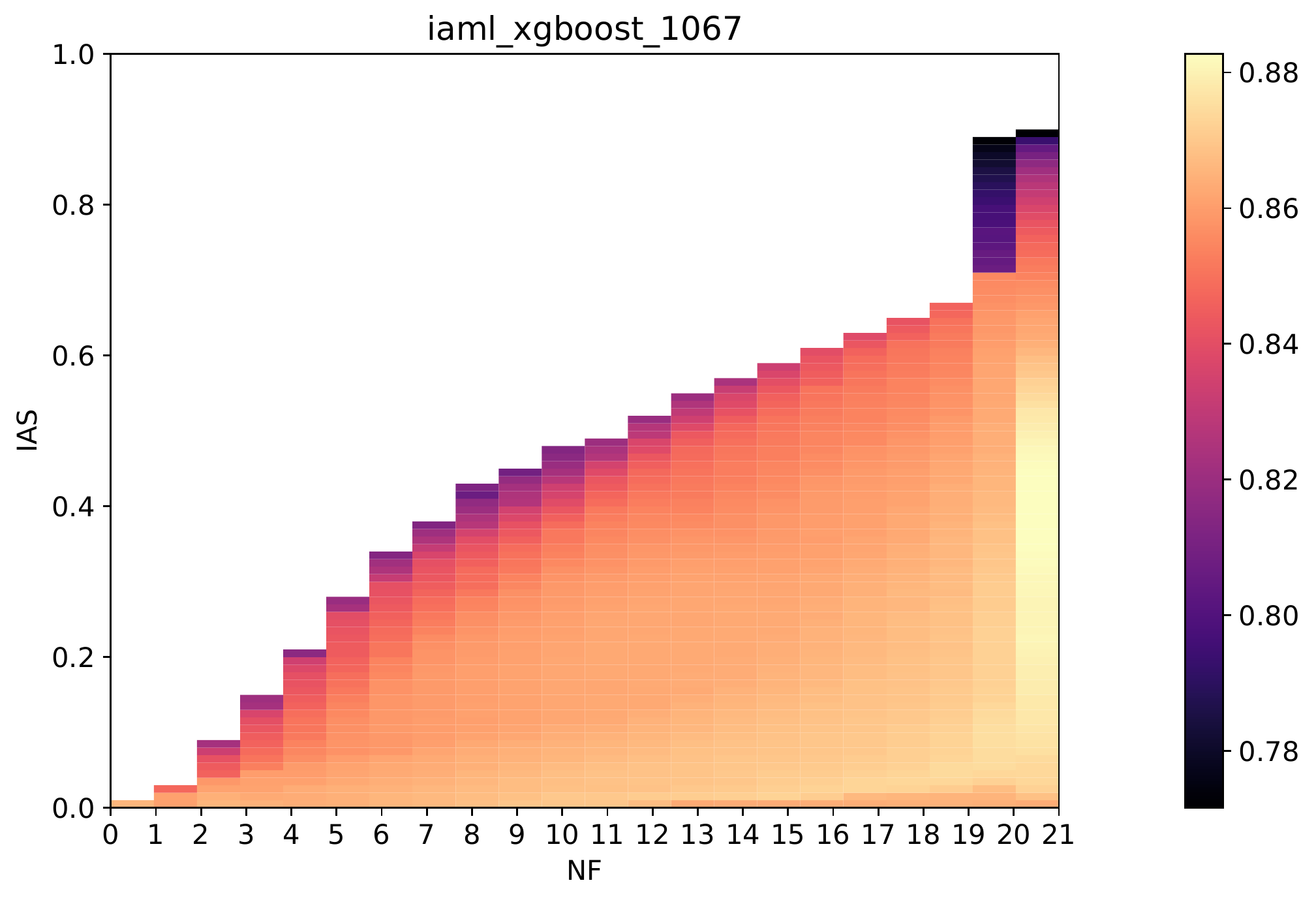}
\endminipage
\caption{Heatmaps of the \texttt{iaml\_xgboost} interpretability benchmark problems.}\label{fig:heatmaps_xgboost_interpretability}
\end{figure}

\begin{figure}[h]
\minipage{0.23\textwidth}
  \includegraphics[width=\linewidth]{Plots/iaml_ranger_rammodel_timepredict_41146.pdf}
\endminipage\hfill
\minipage{0.23\textwidth}%
  \includegraphics[width=\linewidth]{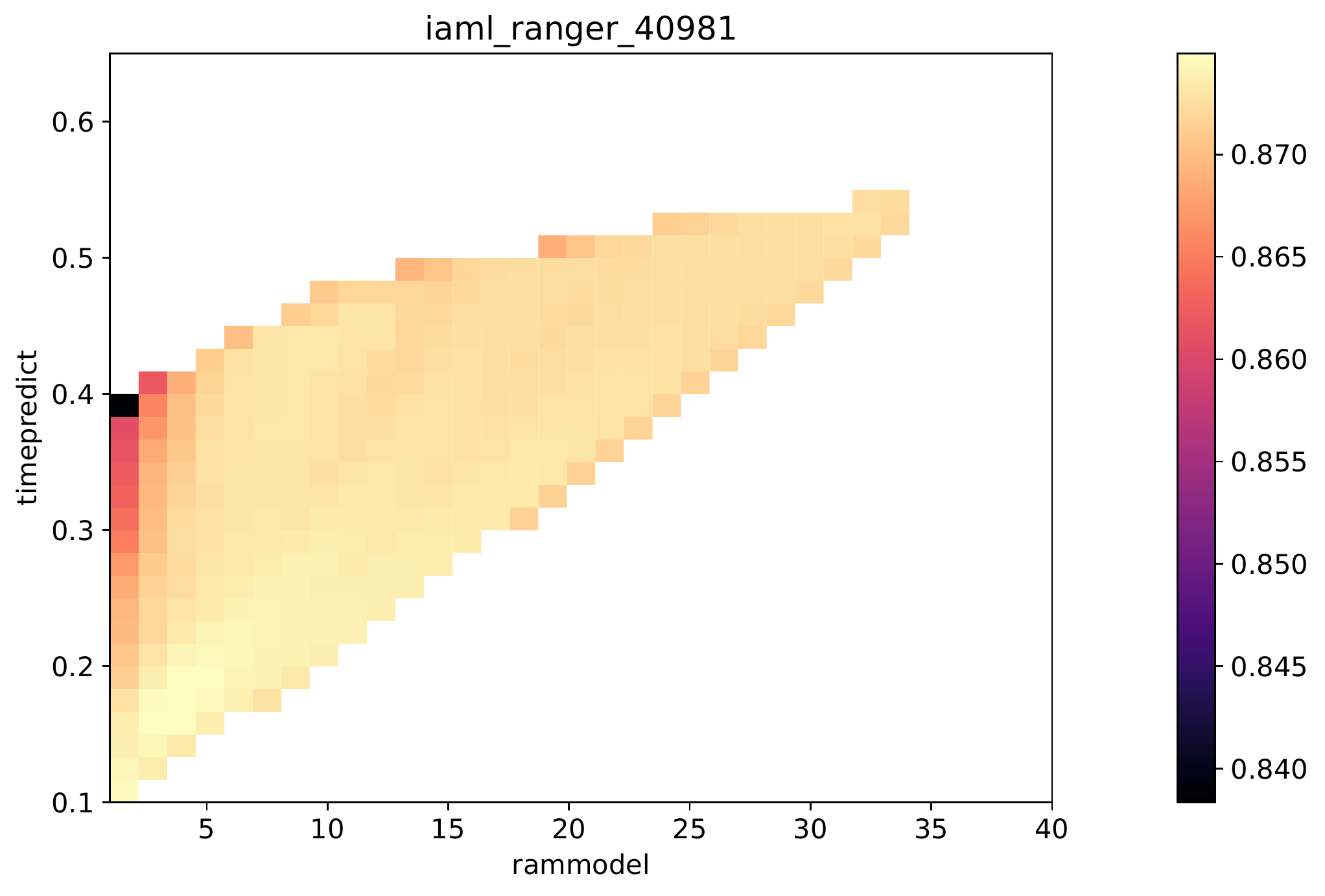}
\endminipage\\
\minipage{0.23\textwidth}
  \includegraphics[width=\linewidth]{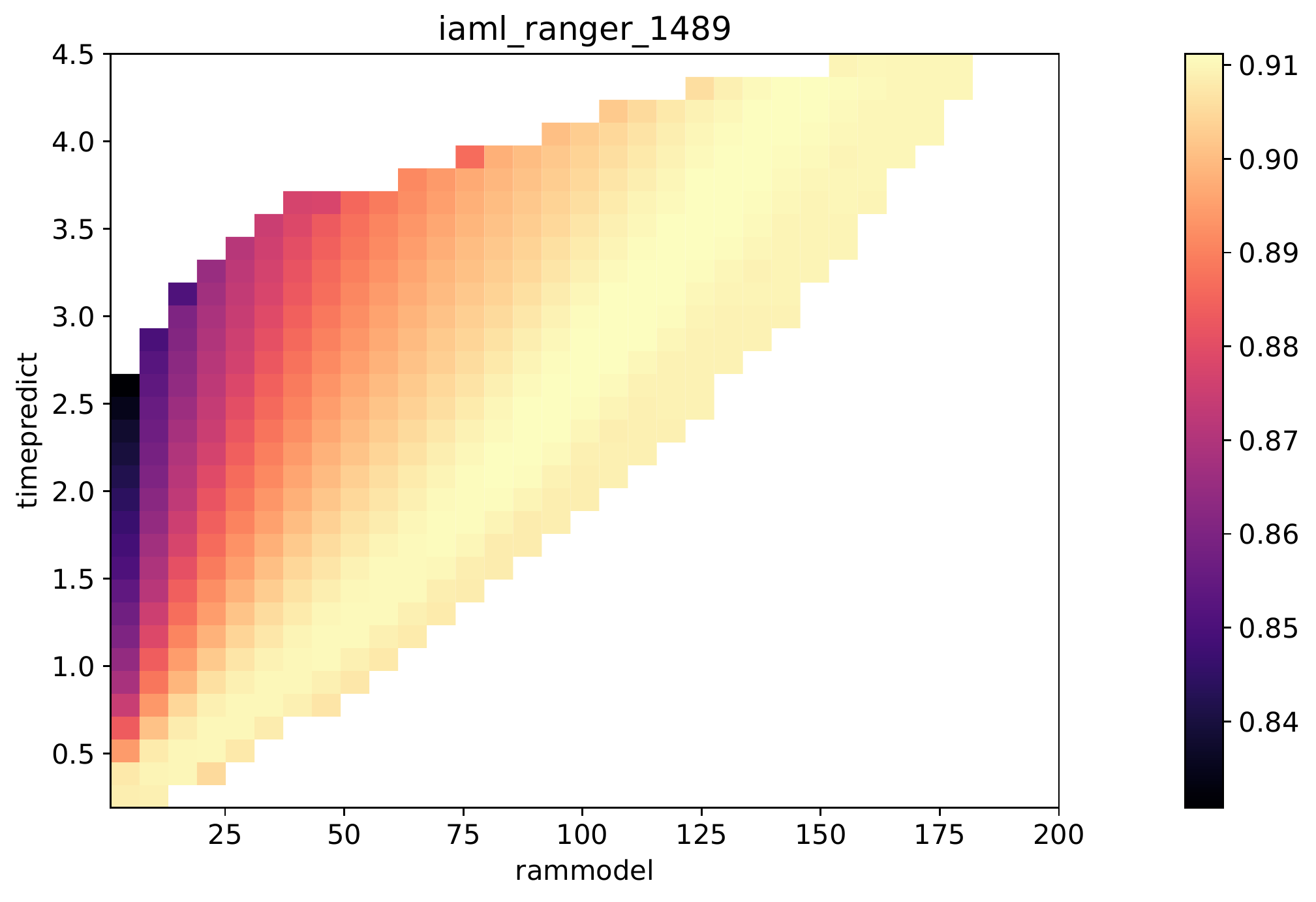}
\endminipage\hfill
\minipage{0.23\textwidth}%
  \includegraphics[width=\linewidth]{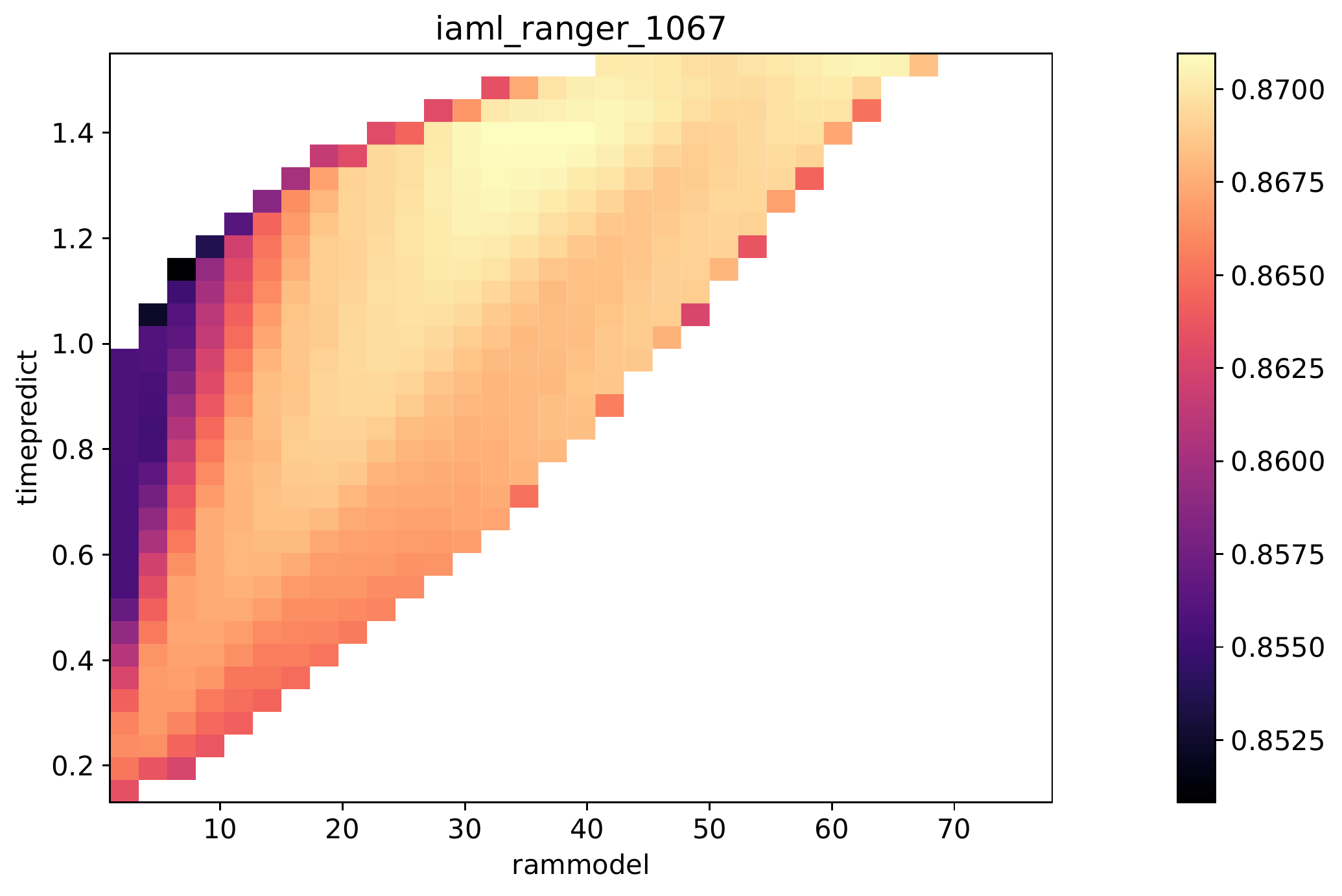}
\endminipage
\caption{Heatmaps of the \texttt{iaml\_ranger} resource usage benchmark problems.}\label{fig:heatmaps_ranger_hardware}
\end{figure}

In this section, we provide heatmaps for all our benchmark problems.
Three Gaussian emitters with a standard deviation of $0.1$, $0.2$ or $0.3$ and one random emitter were run each with a batch size of $25$ for $10000$ iterations resulting in $1000000$ evaluations in total.
The color gradient represents the accuracy (objective function), with black indicating lowest accuracy and bright yellow/white indicating highest accuracy.
\Cref{fig:heatmaps_ranger_interpretability} shows heatmaps for the \texttt{iaml\_ranger} interpretability benchmark problems.
\Cref{fig:heatmaps_xgboost_interpretability} shows heatmaps for the \texttt{iaml\_xgboost} interpretability benchmark problems.
\Cref{fig:heatmaps_ranger_hardware} shows heatmaps for the \texttt{iaml\_ranger} resource usage benchmark problems.

\section{Technical Details}
Benchmark experiments were conducted on a single Intel Core i7-10510U CPU and took around 7 hours in total.
YAHPO Gym v1.0 was used.
For more details on how the \texttt{iaml\_ranger} and \texttt{iaml\_xgboost} benchmark scenarios of YAHPO Gym were constructed, please see \cite{pfisterer2022}.

\end{document}